\documentclass{article}
\usepackage{ijcai16}
\usepackage{mathrsfs}
\usepackage{color}
\usepackage{graphicx}
\usepackage{url}
\usepackage{times}
\usepackage{amsmath}
\usepackage{times}
\title{Makeup like a superstar: Deep Localized Makeup Transfer Network}
\author{Si Liu$^1$, Xinyu Ou$^{1, 2, 3}$, Ruihe Qian$^{1, 4}$, Wei Wang$^1$ and Xiaochun Cao$^1$ \thanks{corresponding author}\\
$^1$State Key Laboratory of Information Security, Institute of Information Engineering, Chinese Academy of Sciences\\
$^2$School of Computer Science and Technology, Huazhong University of Science and Technology\\
$^3$YNGBZX, Yunnan Open University\\
$^4$University of Electronic Science and Technology of China, Yingcai Experimental School\\
\{liusi, caoxiaochun\}@iie.ac.cn, ouxinyu@hust.edu.cn, 406618818@qq.com, wang$\_$wei.buaa@163.com}

\begin{document}

\maketitle

\begin{abstract}
%
In this paper, we  propose a novel Deep Localized Makeup Transfer Network  to automatically recommend the most suitable makeup for a female and synthesis the makeup on her face. Given a before-makeup face, her most suitable makeup is determined automatically. Then, both the before-makeup and the reference faces are {fed} into the  proposed Deep Transfer Network to generate the  after-makeup face. Our end-to-end makeup transfer network have several nice properties including:
\textbf{(1)} with  complete functions: including  foundation, lip gloss, and eye shadow transfer;
\textbf{(2)} cosmetic specific: different cosmetics are transferred in different manners;
%
%
\textbf{(3)} localized: different cosmetics are applied on different facial regions;
\textbf{(4)} producing naturally looking results without obvious artifacts;
\textbf{(5)} controllable makeup lightness: various results from light makeup to heavy makeup can be generated.
Qualitative and quantitative experiments show that our network performs much better  than the methods of \cite{guo2009digital} and two  variants of NerualStyle \cite{gatys2015neural}.
\end{abstract}

\section{Introduction}

Makeup makes the people more attractive, and there are more and more commercial facial makeup systems in the market.
Virtual Hairstyle\footnote{http://www.hairstyles.knowage.info/} provides manual hairstyle try-on.
Virtual Makeover TAAZ\footnote{http://www.taaz.com/} offers to try some pre-prepared cosmetic elements, such as, lipsticks and eye liners.
However, all these softwares  rely on the pre-determined cosmetics which  cannot meet up with users' individual needs.

\begin{figure}[t]
	\begin{center}
		 \includegraphics[width=0.96\linewidth]{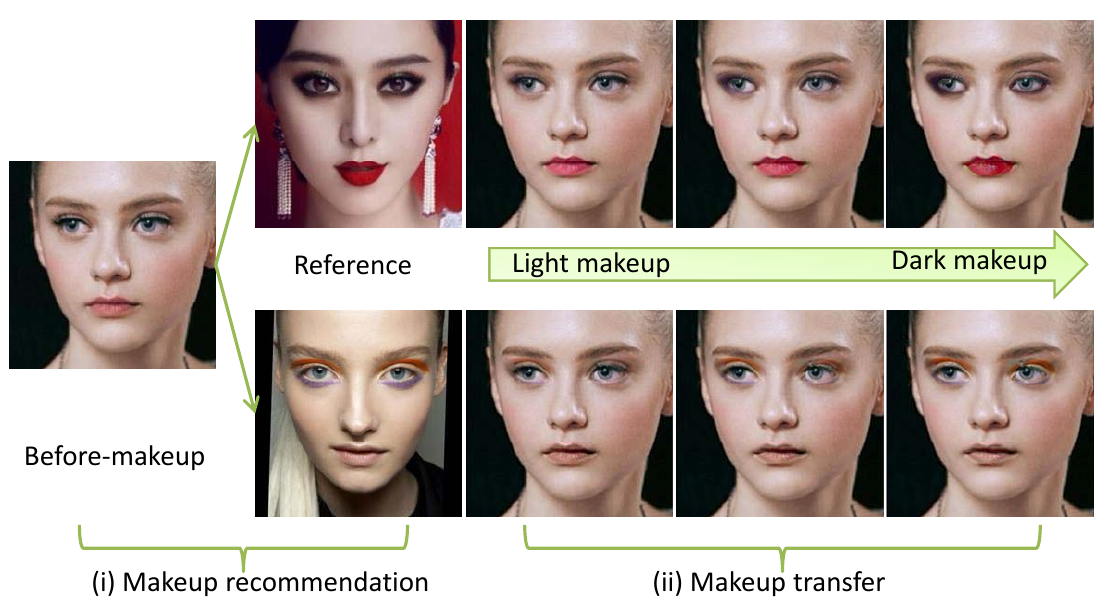}
	\end{center}
		\vspace{-5mm}
	\caption{Our system has two functions. \textbf{\uppercase\expandafter{\romannumeral1}:} recommend the most suitable makeup for each before-makeup face \textbf{\uppercase\expandafter{\romannumeral2}:} transfer the  foundation, eye shadow and lip gloss from the reference to the before-makeup face. The lightness of the makeup can be tuned.
   For better view of all figures in this paper, please refer to the original color pdf file and pay special attention to eye shadow, lip gloss and foundation transfer. }
	\label{fig:first_fig}
		\vspace{-4mm}
\end{figure}

%
Different from the  existing work, our goal is to design a real application system to automatically recommend the most suitable makeup for a female and synthesis the makeup on her face.
As shown in Figure \ref{fig:first_fig}, we simulate an applicable makeup process with two functions.
\textbf{\uppercase\expandafter{\romannumeral1}:} The first function is \emph{makeup recommendation}, where personalization is taken special cares.
More specifically, females with similar face,  eye  or  mouth shapes are suitable for similar makeups \cite{liu2014wow}.
To this end, given a before-makeup face, we find the visually similar face from a reference dataset.
The similarity is measured by the Euclidean distance between deep features produced by an off-the-shelf deep face recognition network \cite{parkhideep}.
To sum up, the recommendation is personalized,  data-driven and easy to implement.
%
\textbf{\uppercase\expandafter{\romannumeral2}:} The second function is \emph{makeup transfer} from the reference face to the before-makeup face.
The makeup transfer function should satisfy five criteria.
1) \emph{With  complete functions}: we consider three kinds of most commonly used cosmetics, i.e., foundation, eye shadow and lip gloss.  Note that our model is quite generic and easily extended to other types of cosmetics.
2) \emph{Cosmetic specific}: different cosmetics are transferred  in their own ways. For example,   maintaining the shape is important for the eye shadow rather than for the foundation.
3) \emph{Localized}: all cosmetics are applied locally on their corresponding facial parts. For example, lip gloss is put on the lip while eye shadow is worn the eyelid.
4) \emph{Naturally looking}: the cosmetics need to be seamlessly fused  with the before-makeup face. In other words, the after-makeup face should look natural.
5) \emph{Controllable  makeup lightness}: we can change  the lightness of each type of cosmetic.

\begin{figure*}[t]
	\begin{center}
		\includegraphics[width=0.8\linewidth]{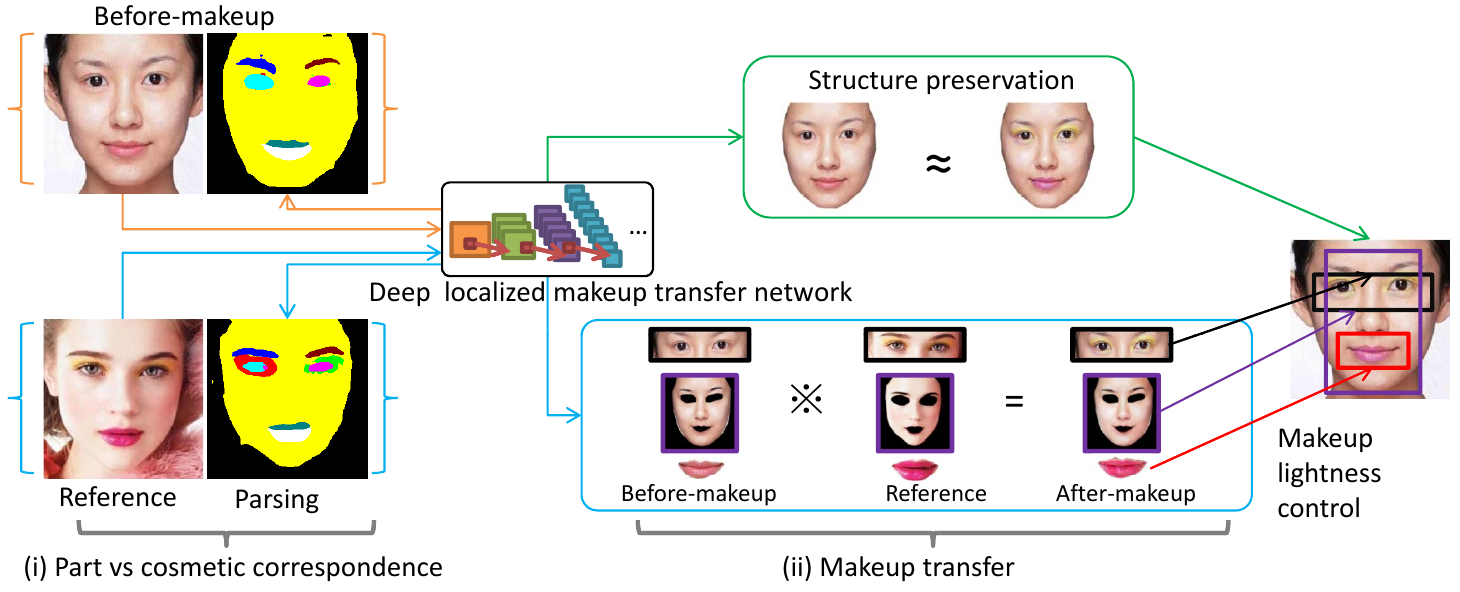}
	\end{center}
		\vspace{-6mm}
	\caption{The proposed Deep Localized Makeup Transfer Network contains two sequential steps.
(i) the correspondences between the facial part (in the before-makeup face) 	and the  cosmetic (in the reference face) are built based on the face parsing network.
(ii)  Eye shadow, foundation and lip gloss are locally transferred with a global smoothness regularization. }
	\label{fig:framework}
    \vspace{-5mm}
\end{figure*}

%
To meet the five afore-mentioned criteria, we propose a Deep Localized Makeup Transfer Network, whose flowchart is shown in Figure \ref{fig:framework}.
The network transfers the makeup from the recommended reference face to a before-makeup face.
Firstly, both before-makeup and reference faces are fed into a face parsing network to generate two corresponding labelmaps.
The parsing network is based on   the Fully Convolutional Networks \cite{long2014fully}  by   (i) emphasizing the makeup relevant facial parts, such as eye shadow and (ii) considering the symmetry  structure of the frontal faces.
Based on the parsing results, the {local} region of the before-makeup face (e.g., mouth) corresponds to its counterpart (e.g., lip) in the reference face.
%
%
Secondly, three most common cosmetics, i.e, eye shadow, lip gloss and foundation are transferred in their own manners.
%
%
For example, keeping the shape is the most important for
transferring eye shadow while the smoothing the skin's texture is the most important for foundation.
So the eye shadow is transferred via directly altering
the corresponding deep features \cite{mahendran2014understanding}  while foundation is transferred via regularizing the inner product of the feature maps \cite{gatys2015neural}.
The after-makeup face is initialized as the before-makeup face, then gradually updated via Stochastic Gradient Descent to produce {naturally looking} results.
%
By tuning up the weight of each cosmetic, a series of after-makeup faces with increasingly heavier makeup can be generated.
%
%
In this way, our system  can produce various results with {controllable makeup lightness}.

Compared with traditional makeup transfer methods \cite{guo2009digital,tong2007example,scherbaum2011computer,liu2014wow}, which   require complex data  preprocessing or annotation and their results are inferior, our contributions are as follows.
%
%
\textbf{\uppercase\expandafter{\romannumeral1}:}  To the best of our knowledge, it is the first makeup transferring method based on deep learning framework and can produce very natural-looking results.
Our system can transfer foundation, eye shadow and lip gloss, and therefore is with complete functions.
Furthermore, the lightness of the makeup can be controlled to meet the needs of various users.
%
\textbf{\uppercase\expandafter{\romannumeral2}:} we propose an end-to-end Deep Localized Makeup Transfer Network to first build part vs cosmetic correspondence and then transfer makeup.
%
Compared with NerualStyle \cite{gatys2015neural} which fuses two images globally, our method transfers makeup locally from the cosmetic regions to their corresponding facial parts. Therefore, a lot of unnatural results are avoided.

\vspace{-2mm}

\section{Related Work}
We will introduce the related makeup transfer methods and the most representative deep image synthesis methods.

\textbf{Facial Makeup Studies:} The work of Guo \emph{et al.}  \cite{guo2009digital} is the first attempt for the makeup transfer task.
It first  decomposes the before-makeup and reference faces into three layers.
Then, they transfer information between the corresponding layers.
One major disadvantage is that it needs warp the reference face to the before-makeup face, which is very challenging.
Scherbaum \emph{et al. } \cite{scherbaum2011computer} propose  to use a 3D morphable face model to facilitate facial makeup.
%
It also requires  the before-after makeup face pairs for the same person, which are difficult to collect in real application.
Tong  \emph{et al. } \cite{tong2007example}  propose a  ``cosmetic-transfer" procedure to realistically transfer the cosmetic style captured in the example-pair to another person's face.
%
%
The requirement of before-after  makeup pairs limits the practicability  of the system.
Liu et al \cite{liu2014wow} propose an automatic makeup recommendation and synthesis system called beauty e-expert.
Their contribution is in the recommendation module.
To sum up, our method greatly relaxes the requirements of traditional makeup methods, and generates more naturally-looking results.

\textbf{Deep Image Synthesis Methods:}  Recently,  deep learning has achieved great success in
fashion analysis works \cite{liu2015matching,liang2015deep}.
Dosovitskiy \emph{et al.} \cite{dosovitskiy2014learning,dosovitskiy2015inverting} use a generative CNN  to  generate images of objects given object type, viewpoint, and color.
Simonyan \emph{et al.} \cite{simonyan2013deep} generate an image, which  visualizes the notion of the class captured by a net.
Mahendran \emph{et al.} \cite{mahendran2014understanding} contribute a general framework to invert both hand-crafted and deep representations to the images.
Gatys \emph{et al.}  \cite{Gatys2015b} present  a parametric texture model based on the CNN which can synthesise high-quality natural textures.
Generative adversarial network  \cite{goodfellow2014generative}   consists of two components; a generator and a discriminator. The generated image is very natural without obvious artifacts.
Goodfellow \emph{et al.} introduce \cite{Goodfellow}  a simple and fast method of generating adversarial examples. Their main aim is  to enhance the CNN training instead image synthesis.
Kulkarni \emph{et al.} \cite{kulkarni2015deep}   present the Deep Convolution Inverse Graphics Network, which learns an interpretable representation of images for 3D rendering.
All existing deep methods can only generate one image. However, we mainly focus on how to generate a new image having  the nature of the two input images.
Recently, deep learning techniques have been applied in many image editing tasks, such as image colorization \cite{deepColorization}, photo adjustment  \cite{yan2014automatic}, filter learning \cite{xu2015deep},  image inpainting \cite{ren2015shepard}, shadow removal \cite{shen2015shadow} and  super-resolution \cite{dong2014learning}.
These methods are operated on a single image.
%
NerualStyle \cite{gatys2015neural} is the most similar with us. They use CNN to synthesis a new image by combining  the structure layer from one image and the style layers of another.
The key difference between their work and ours is that our network is applied locally, which can produce more natural results.

%

\vspace{-2mm}

\section{Approach}
In this Section, we sequentially introduce our makeup recommendation and makeup transfer methods in detail.


\vspace{-2mm}

\subsection{Makeup Recommendation}
The most important criterion of makeup recommendation is \emph{personalization} \cite{liu2014wow}.

\begin{figure}[h]
	\begin{center}
		 \includegraphics[width=0.95\linewidth]{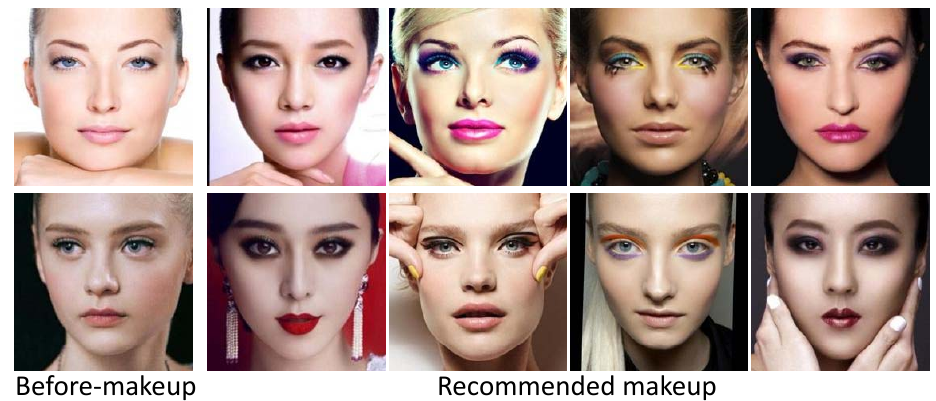}
	\end{center}
		\vspace{-6mm}
	\caption{Two examples of makeup recommendation. The first columns are the before-makeup faces, other columns are the recommended reference faces. }
	\label{fig:recommendation}
			\vspace{-4mm}
\end{figure}

Females that look like are suitable for similar makeups.
Given a before-makeup face, we find several similar faces in the reference dataset.
The similarities are defined as the Euclidean distances between the deep features extracted by feeding the face into an off-the-shelf face recognition model \cite{parkhideep} named VGG-Face.
The deep feature is the concatenation of the  two ${\ell _2}$ normalized   FC-$4096$ layers.
The VGG-Face model is trained based on  VGG-Very-Deep-$16$ CNN architecture \cite{simonyan2014very} and aims to accurately identify different people regardless of whether she puts on makeup, which meets our requirement.
Therefore,  the extracted features can capture the most discriminative structure of faces.
%
Finally, the retrieved results serve as the reference faces to transfer their cosmetics to the before-makeup face.
Figure \ref{fig:recommendation} shows the makeup recommendation results.
It illustrates  that the recommended reference faces have similar facial shapes with the before-makeup faces, and therefore the recommendation is personalized.




\vspace{-2mm}

\subsection{Facial Parts vs. Cosmetics Correspondence} \label{sec:face_parsing}
In order to transfer the makeup, we need build the correspondence between facial parts of the  before-makeup face and the cosmetic regions of the reference face.
As a result, the cosmetic can be between the matched pairs.
Most of the correspondences can be obtained by the face parsing results, e.g., ``lip'' vs ``lip gloss''.
The  only exception is the eye shadow transfer. Because the before-makeup face does not have eye shadow region and the shape of the eyes are different, we need to warp the eye shadow shape of the reference face.

%
%

\textbf{Face Parsing:} Our face parsing model is based on the Fully Convolution Network (FCN) \cite{long2014fully}.
%
It is trained using both the before-makeup and reference faces.
The $11$ parsing labels are shown in Table \ref{tab:labelmap}.  The network  takes input of arbitrary size and produces correspondingly-sized output with efficient inference and learning.

\begin{table*}[t]
	\begin{tabular}{|c|c|c|c|c|c|c|c|c|c|c|}
		\hline
		\hspace{-2mm}		Labels \hspace{-2mm} &		\hspace{-2mm}		L/R eye \hspace{-2mm} & L/R eyebrow \hspace{-2mm} & inner mouth \hspace{-2mm}   	  &		\hspace{-2mm}		L/R eye shadow \hspace{-2mm} & Up/Low lip (lip gloss) \hspace{-2mm} &  \hspace{-2mm} background   		 \hspace{-2mm}			&	Face (Foundation)	   \\
		\hline
		\hspace{-2mm} Subsets \hspace{-2mm} & 		\hspace{-2mm}	both \hspace{-2mm} & both \hspace{-2mm} & both \hspace{-2mm}  & 		ref	 \hspace{-2mm} & before (ref) \hspace{-2mm} & both \hspace{-2mm}   & 		 \hspace{-2mm}  before (ref) 	 \\
		\hline
	\end{tabular}
		\vspace{-2mm}
	\caption{ $11$ Labels from both before-makeup (referred to as ``before'') and reference face sets (referred to as ``ref''). Certain labels, such as ``L eye'',  belong to both dataset,  while certain labels such as ``L eye shadow''  belong  to reference face set only.
		``L'',  ``R'', ``Up'' and ``Low'' stands for ``Left'', ``Right'', ``Upper'' and ``Lower'' respectively. }
	\label{tab:labelmap}
	\vspace{-6mm}
\end{table*}

When training the face parsing model, we pay more attention to the makeup relevant labels.
For example, compared with ``background'',  we bias toward the  ``left eye shadow''.
Therefore, we propose a \emph{weighted cross-entropy loss} which is a weighted sum over the spatial dimensions of the final layer:

\vspace{-3mm}

\begin{equation}
\ell \left( {x;\theta } \right) = \sum\nolimits_{ij} {\ell '\left( {{{\rm{y}}_{ij}},p\left( {{x_{ij}};\theta } \right)} \right) \cdot w\left( {{y_{ij}}} \right)},
\end{equation}
where $\ell '$ is the cross entropy loss
defined on each pixel.
${{y_{ij}}}$  and ${p\left( {{x_{ij}};\theta } \right)}$
are the ground truth and predicted label of the pixel  ${{x_{ij}}}$, 	
and  ${w\left( {{y_{ij}}} \right)}$ is the label weight.
The weight is set empirically by maximizing  the F1 score in the validation set.

\begin{figure}[h]
	\begin{center}	
		\includegraphics[width=0.85\linewidth]{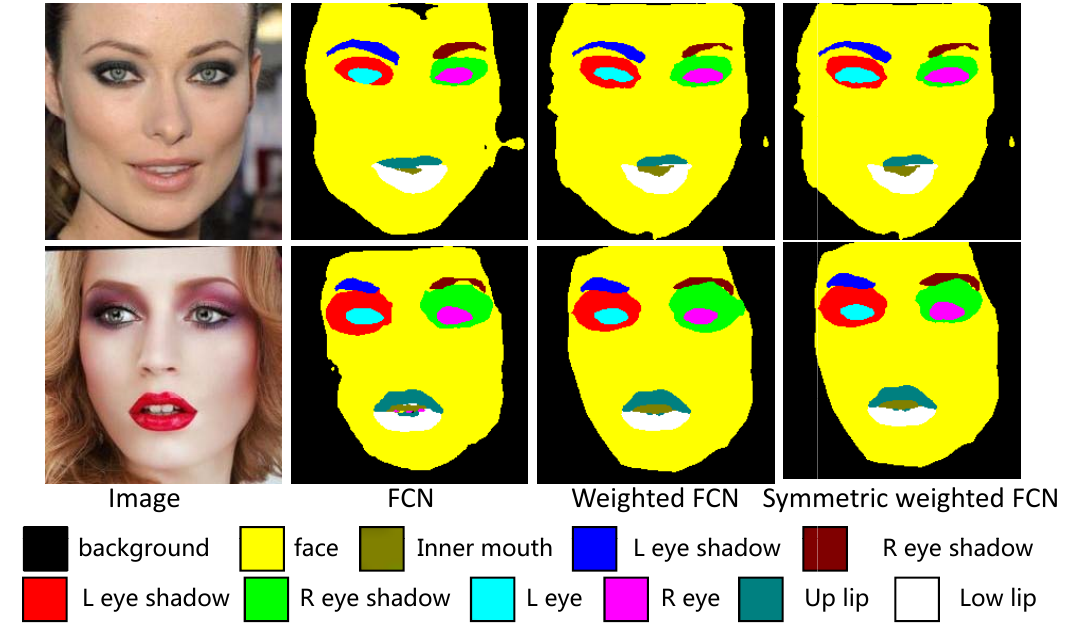}	
	\end{center}
		\vspace{-4mm}
	\caption{Two face parsing results. The input image, the results parsed by FCN, weighted FCN and symmetric weighted FCN are shown sequentially. }
	\label{fig:face_detection}
	\vspace{-4mm}
\end{figure}

Because  the faces in the collected datasets are in frontal view,
and preprocessed by face detection and  alignment\footnote{www.faceplusplus.com/}.
In the testing phase, we enforce the symmetric prior and  replace the prediction confidence of
both point $p$ and its horizontally mirrored  counterparts ${f\left( p \right)}$
by their average  ${x_{p,c}} = \frac{1}{2}\sum {\left( {{x_{p,c}} + {x_{f\left( p \right),c}}} \right)}$,
where $c$ denotes the channels.
Like  \cite{chen14semantic,papandreou15weak}, the  symmetric prior is only added in the testing phase currently.
In the further we will explore how to enforce the structure prior in the training phase.
Figure \ref{fig:face_detection} shows the parsing results of the original FCN, weighted FCN, and symmetric weighted FCN.


%

		

\textbf{Eye Shadow Warping}:
%
Based on the parsing results, most correspondences, e.g., ``face'' vs ``foundation'', are built.
However, there is no eye shadow in the before-makeup face,
therefore we need to generate an eye shadow mask in the before-makeup face.
Moreover, because the shapes of eyes and eye brows are different in the face pair,
the shape of eye shadow need to be slightly warped.
More specifically,  we get $8$ landmarks on eyes and eye brow regions,
including inner, upper-middle, lower-middle and  outer corner of eyes and eye brows.
Then the eye shadows are warped by the thin plate spline  \cite{bookstein1989principal}.

\subsection{Makeup Transfer}
Makeup transfer is conducted based on the correspondences among image pairs.
Next we will elaborate how to transfer eye shadow, lip gloss and foundation.
Keeping  the facial structure should also be considered and incorporated into the final objective function.

\begin{figure}[h]
	\begin{center}
		\includegraphics[width=0.95\linewidth]{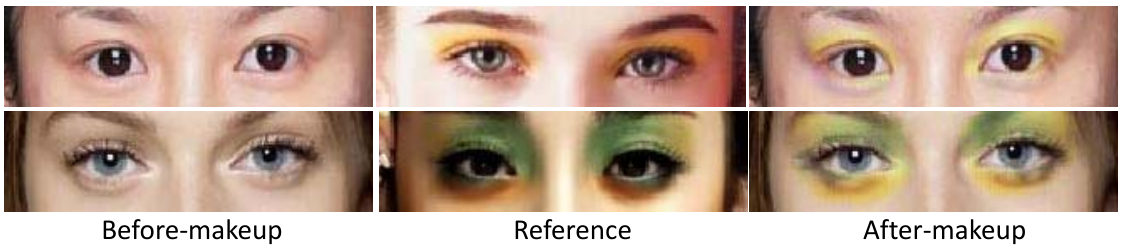}
	\end{center}
    \vspace{-4mm}
	\caption{Two  results of  eye shadow transfer. The before-makeup, reference and after-makeup eye shadow are shown. }
	\label{fig:eye_shadow}
   \vspace{-4mm}
\end{figure}

\textbf{Eye Shadow Transfer} needs to consider both the shape and color.
Let take the left eye shadow as an example. The right eye shadow is transferred similarly.
Suppose  $s_r$ is the binary mask of left eye shadow in the reference face while
$s_b$ is the warped binary mask in the  before-makeup face.
Note that after eye shadow warping, $s_r$ and $s_b$ are of the same shape and size.
Technically, eye shadow transfer is to replace $s_b$'s  deep feature representation in a certain layer (conv1-1 in this paper)  with $s_r$. The mathematical form for the loss of left eye shadow transfer is ${R_{l}}(A)$:

	\vspace{-2mm}
	
\begin{equation}
	\begin{array}{l}
		{A^*} = \mathop {\arg \min }\limits_{A \in {R^{H \times W \times C}}} {R_{l}}(A)\\
\; \; \quad		= \mathop {\arg \min }\limits_{A \in {R^{H \times W \times C}}} \left\| {P\left( {{\Omega ^l}\left( {A\left( {{{\rm{s'}}_b}} \right)} \right)} \right) - P\left( {{\Omega ^l}\left( {R\left( {{{\rm{s'}}_r}} \right)} \right)} \right)} \right\|_2^2
	\end{array}
	\label{eq:eye_shadow}
\end{equation}
where  $H$, $W$ and $C$ are the height, width and channel number of the input image. ${{\Omega ^l}} :{R^{H \times W \times C}} \to {R^d}$ is the $d$-dim feature representation of the conv1-1 layer of the face parsing model, ${A}$ and ${R}$ are the after-makeup face and reference face, respectively.
${{\rm{s'}}_b}$ and ${{\rm{s'}}_r}$ are achieved by  mapping $s_b$ to $s_r$  from  the data layer to the conv1-1 layer via the convolutional feature masking \cite{dai2014convolutional,he2014spatial}.
${A\left( {{{\rm{s'}}_b}} \right)}$ and ${R\left( {{{\rm{s'}}_r}} \right)}$ are the  eye shadow regions corresponding to the masks ${\rm{s'}}_b$ and ${\rm{s'}}_r$.
Similarly, we can define the loss function for right eye shadow ${R_{r}}(A)$.
The results for both eye shadow transfer are shown in Figure \ref{fig:eye_shadow}, where both the color and shape of the eye shadows are transferred.

%

\vspace{-2mm}

\begin{figure}[h]
	\begin{center}
		\includegraphics[width=0.7\linewidth]{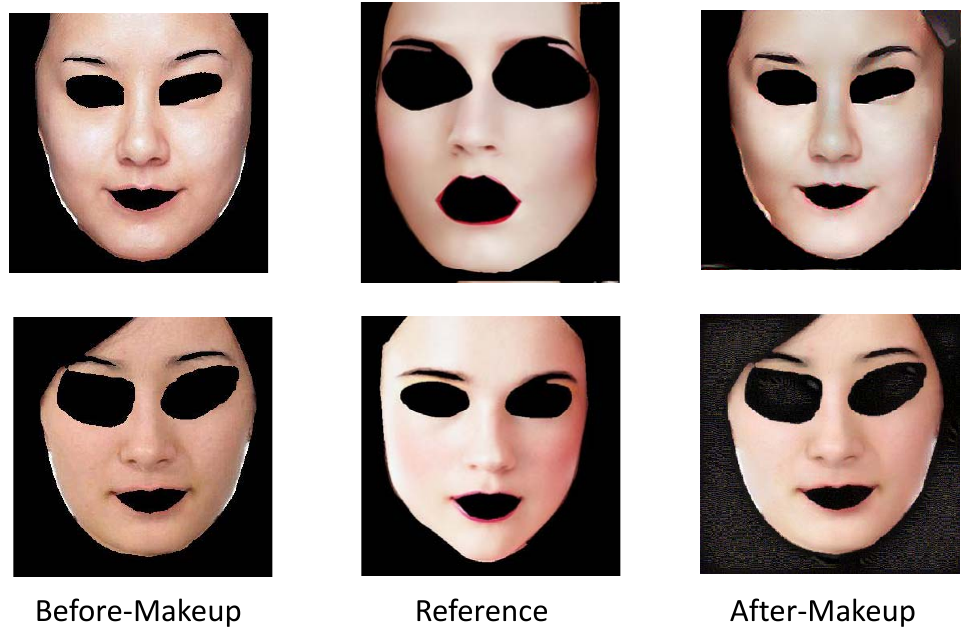}
	\end{center}
	\vspace{-6mm}
	\caption{Two exemplars of  foundation transfer. In each row, before-makeup, reference and after-makeup are shown. }
	\label{fig:foundation}
		\vspace{-3mm}
\end{figure}

\textbf{Lip Gloss and Foundation Transfer} require transferring color and texture.  The lip gloss  ${R_{f}}(A)$ is defined as in \eqref{equ:foundation}.

\begin{equation}
	\begin{array}{l}
		{A^*} = \mathop {\arg \min }\limits_{A \in {R^{H \times W \times C}}} {R_f}(A)\\
		\quad \; \; = \mathop {\arg \min }\limits_{A \in {R^{H \times W \times C}}} \sum\limits_{l = 1}^L {\left\| {\Omega _{ij}^l\left( {A\left( {{{\rm{s'}}_b}} \right)} \right) - \Omega _{ij}^l\left( {R\left( {{{\rm{s'}}_r}} \right)} \right)} \right\|_2^2}
	\end{array}
	\label{equ:foundation}
\end{equation}
Here, $L$ is the number of layers used.
Technically, we use $5$ layers, including conv1-1, conv2-1, conv3-1, conv4-1 and conv5-1. The Gram matrix ${\Omega ^l} \in {R^{{N_l} \times {N_l}}}$ is defined in \eqref{equ:grammatrix}, where ${{N_l}}$ is the number of feature maps in the $l$-th layer and $\Omega _{ij}^l$ is the inner product between the vectorised feature map $i$ and $j$ in layer $l$:
\begin{equation}
	\Omega _{ij}^l = \sum\limits_k {\Omega _{ik}^l\Omega _{jk}^l}
    \label{equ:grammatrix}
\end{equation}
The results for foundation transfer are shown in Figure \eqref{fig:foundation}. The girl's skin is exquisite after the foundation transfer.

\vspace{-2mm}

\begin{figure}[h]
	\begin{center}
		\includegraphics[width=0.6\linewidth]{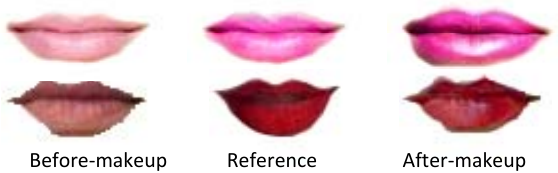}
	\end{center}
	\vspace{-6mm}
	\caption{Two exemplars of lip gloss transfer results.  In each row, before-makeup, reference and after-makeup are shown. }
	\label{fig:lip}
		\vspace{-3mm}
\end{figure}

The upper lip gloss loss ${R_{up}}(A)$ and lower lip gloss loss ${R_{low}}(A)$ are defined in similar way as \eqref{equ:foundation}. And the lip gloss transfer results are shown in Figure \ref{fig:lip}. After lip gloss transfer, the colors of the lip are changed to the reference lip.

\vspace{-2mm}

\begin{figure}[t]
	\begin{center}
		 \includegraphics[width=0.8\linewidth]{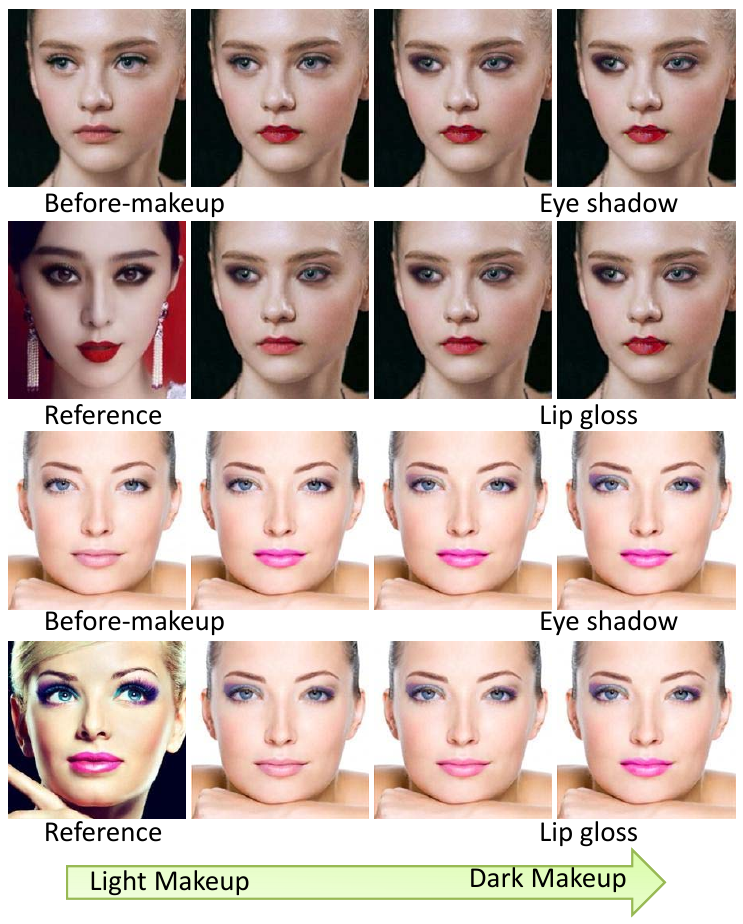}
	\end{center}
	\vspace{-6mm}
	\caption{The lightness of the makeup can be controlled.	}
	\label{fig:lightness}
	\vspace{-4mm}
\end{figure}

\textbf{Structure Preservation} term ${R_s}(A)$ is defined as in \eqref{eq:eye_shadow}.
The only difference is that every elements of $s_b$ and  $s_r$ are $1$.

\textbf{Overall Makeup Transfer}  considers eye shadow, lip gloss and foundation, and also preserves the face structure.
\vspace{-2mm}
\begin{equation}
\begin{array}{l}
{A^*} = \mathop {\arg \min }\limits_{A \in {R^{H \times W \times C}}} {\lambda _e}\left( {{R_{l}}(A) + {R_{r}}(A)} \right) + {\lambda _f}{R_f}(A) \\
 \quad  \quad  \quad + {\lambda _l}\left( {{R_{up}}(A) + {R_{low}}(A)} \right) +  {\lambda _s}{R_s}(A) + {R_{{V^\beta }}}(A)
\end{array}
	\label{equ:fusion}
\end{equation}
To make the results more natural,  the total variance term
${R_{{V^\beta }}}{\rm{ = }}\sum\nolimits_{i,j} {{{\left( {{{\left( {{A_{i,j + 1}} - {A_{ij}}} \right)}^2}{\rm{ + }}{{\left( {{A_{i{\rm{ + }}1,j}} - {A_{ij}}} \right)}^2}} \right)}^{\frac{\beta }{2}}}}$ is added.    ${R_{l}}(A)$,  ${R_{r}}(A)$, ${R_f}(A)$, ${R_{up}}(A)$, ${R_{low}}(A)$  and ${R_{s}}(A)$
are the left, right eye shadow, foundation, upper, lower lip gloss and face structure loss. And  ${\lambda _e}$, ${\lambda _f}$, ${\lambda _l}$ and ${\lambda _e}$ are the weights to balance different cosmetics.  By tuning these weights, the lightness of makeup can be adjusted. For example, by increasing the ${\lambda _e}$, the eye shadow will be darker.


%

%
%

The overall energy function (\ref{equ:fusion}) is optimized via Stochastic  Gradient Descent (SGD)  by using momentum \cite{mahendran2014understanding}:

\vspace{-2mm}

\begin{equation}
	{\mu _{t + 1}} \leftarrow m{\mu _t} - {\eta _t}\nabla E(A) \;\;\;
	{A_{t + 1}} \leftarrow {A_t} + {\mu _t}
\end{equation}
where the  ${\mu _t}$ is  a weighed average of the last several gradients, with decaying factor $m=0.9$. ${A_0}$  is initialized as the before-makeup face.

	\vspace{-4mm}
	
\section{Experiments}

\subsection{Experimental Setting}
%
%
\textbf{Data Collection and Parameters:} We collect a new dataset with $1000$ before-makeup faces and  $1000$  reference faces.
Some before-makeup faces are nude makeup or very light makeup.
Among the $2000$ faces, $100$ before-makeup faces and
$500$ reference faces are randomly selected as test set.
The remaining $1300$ faces and $100$ faces are used as training and validation set.
Given one  before-makeup test face, the most similar ones among the $500$  reference test faces are chosen to transfer the makeup.
The weights $[{\lambda _s} \; {\lambda _e} \;  {\lambda _l} \;  {\lambda _f} ]$ are  set as $[10 \; 40 \; 500 \; 100]$.
%
The weights of different labels in the  weighted FCN are set as  $[1.4 \; 1.2 \; 1]$ for  \{eyebrows, eyes, eye shadows\}, \{lip, inner mouth\} and \{face, background\}, respectively.
These weights are set empirically by the validation set.

\textbf{Baseline Methods:} To the best of our knowledge, the only makeup transfer work is  Guo and Sim \cite{guo2009digital}. %
We also compare with two variants of Gatys \emph{et al.}'  method \cite{gatys2015neural}.
They use CNN to synthesis a new image by combining  the content layer  from one image and the style layers of another image.
The first variant is  called NerualStyle-CC treating  both before-makeup and reference faces as content.
Another variant is named as NerualStyle-CS which uses the before-makeup as content and reference face as style.
We cannot compare with other related makeup methods, such as, Tong et al. \cite{tong2007example}, Scherbaum et al. \cite{scherbaum2011computer} or Beauty E-expert system \cite{liu2014wow} require before-and-after makeup image pairs, 3D information or extensive labeling of facial attributes.
%
%
%
%
%
%
The proposed model can transfer the makeup in $6$ seconds for an $224 \times 224$ image pair using TITAN X GPU.

%
%
%

%
%
%

 \begin{figure}[t]
 	\begin{center}		 \includegraphics[width=1\linewidth]{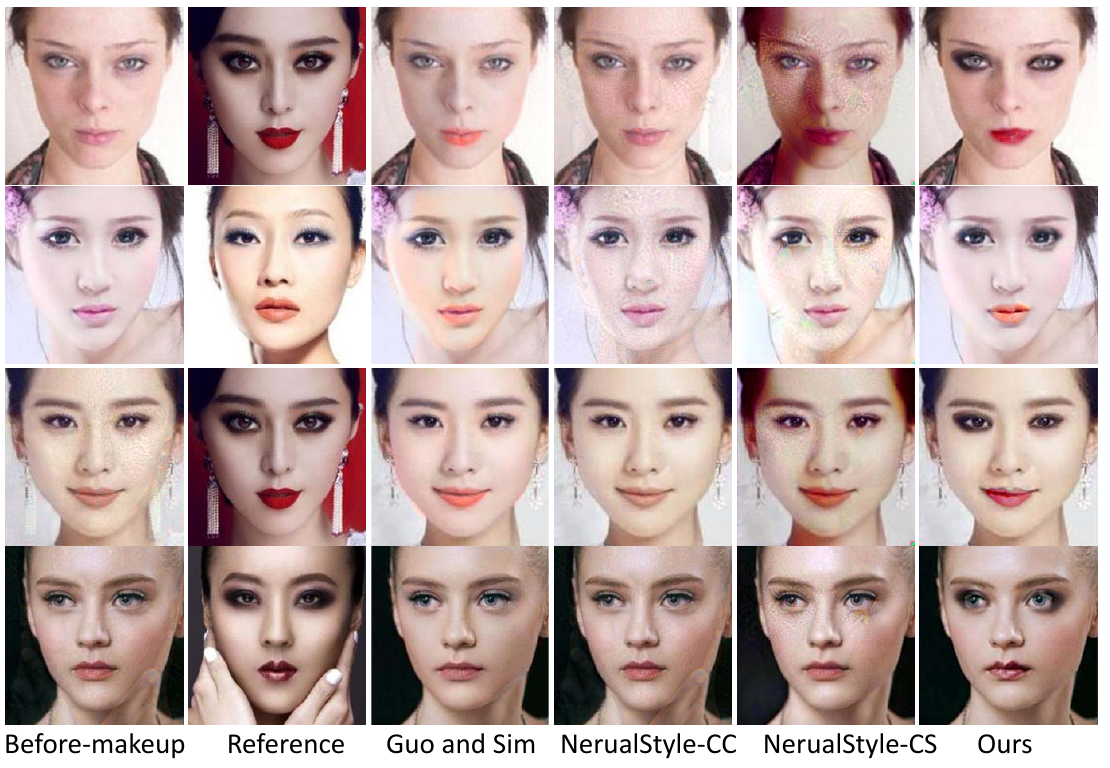}
 	\end{center}
 	\vspace{-5mm}
 	\caption{Qualitative comparisons between the state-of-the-arts and ours.	}
 	\label{fig:baseline}
 	 	\vspace{-5mm}
 \end{figure}

 \begin{figure*}[t]
 	\begin{center}
 		 \includegraphics[width=0.93\linewidth]{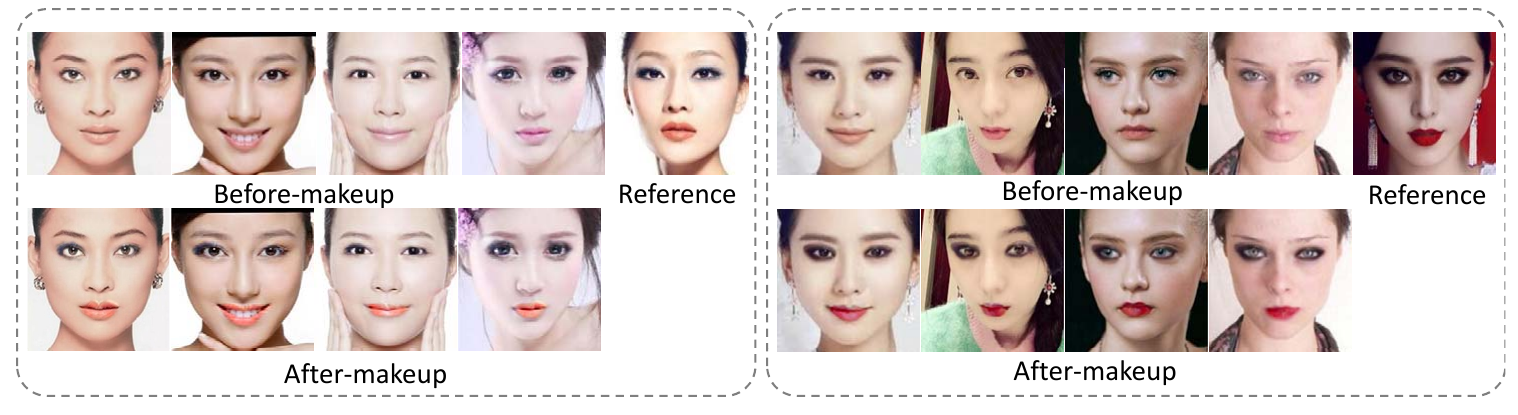}
 	\end{center}
 	\vspace{-5mm}
 	\caption{Different girls wear the same makeup.	}
 	\label{fig:more_result_same_reference}
 	\vspace{-4mm}
 \end{figure*}

 \begin{figure*}[t]
 	\begin{center}
 		 \includegraphics[width=0.93\linewidth]{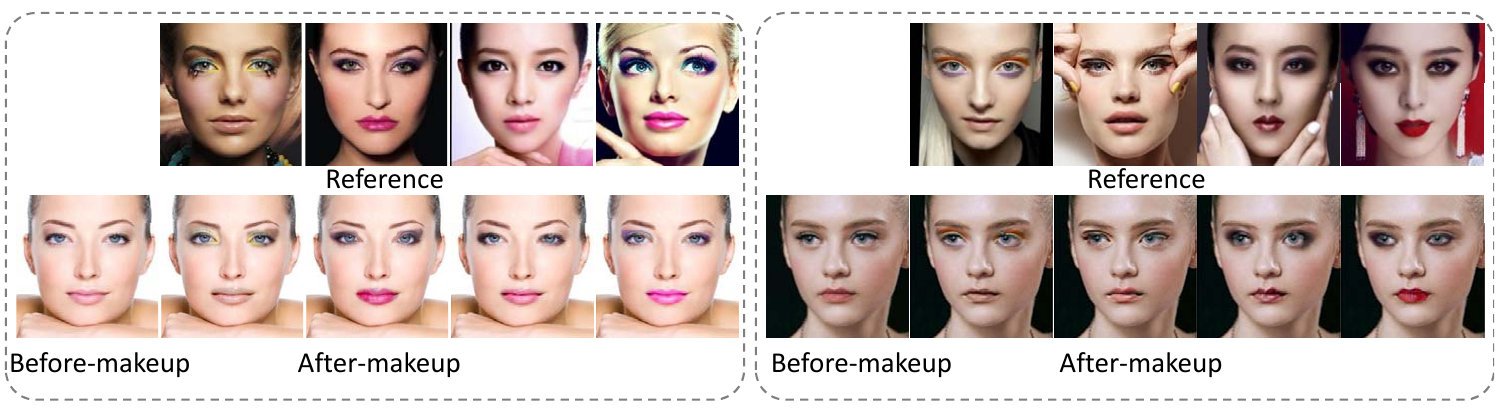}
 	\end{center}
 	\vspace{-5mm}
 	\caption{The same girl wears different makeups. }
 	\label{fig:more_result_same_girl}
 	\vspace{-6mm}
 \end{figure*}

%
%
%
%

\subsection{Makeup Lightness}
In order to show our method can generate after-makeup face with various makeup lightness, ranging from light makeup to dark makeup, we gradually increase certain makeup weights ${\lambda _e}$, ${\lambda _f}$ and ${\lambda _l}$.
Four results are shown in  Figure \ref{fig:lightness}.
The first two rows use the same before-makeup and reference faces.
The girl's eye shadows become  gradually darker in the first row.
In the second row, the lip color becomes redder while other cosmetics keep unchanged.
The third and fourth rows show another example.
The eye shadow and lip gloss are increasingly darker in the third and last row, respectively.

\subsection{Comparison with State-of-the-arts}
%
%

We compare with Guo and Sim\footnote{We sincerely thank the authors for sharing their code.} \cite{guo2009digital}, NerualStyle-CC and NerualStyle-CS\footnote{we use the code https://github.com/jcjohnson/neural-style and fine-tune the parameter for best visual effects.} \cite{gatys2015neural}.
Follwing the evaluation metric in \cite{liu2014wow}, the comparison is conducted  both qualitatively and  quantitatively.

The \textbf{qualitative} results are shown in Figure \ref{fig:baseline}.
Both Guo and Sim \cite{guo2009digital} and ours produce naturally looking results.
However, our result  transfers the lip gloss and eye shadows with the same lightness as the reference faces, while
Guo and Sim \cite{guo2009digital} always transfers  much lighter makeup than the reference face.
For example, in the first row, the lip gloss of both the reference face and our results are deep red. But the lip gloss of Guo and Sim is orange-red.
In addition, the eye shadows of both the reference face and our results are very dark and smoky. However, Guo and Sim can only produce very light eye shadow.
Compared with NerualStyle-CC and NerualStyle-CS \cite{gatys2015neural}, our after-makeup faces contain  much less artifacts.
%
It is because our makeup transfer is conducted between the local regions, such as lip vs lip gloss while the NerualStyle-CC and NerualStyle-CS \cite{gatys2015neural} transfer the makeup globally. Global makeup transfer suffers from the mismatch problem between the image pair.
It shows the advantages of our deep localized makeup transfer network.

The \textbf{quantitative} comparison mainly focuses on the quality of makeup transfer and the degree of harmony.
%
For each of  the $100$ before-makeup test faces,  five most similar reference faces are found.
Thus we have totally $100 \times 5$ after-makeup results for each makeup transfer method.
We compare our method with each of the $3$ baselines sequentially.
Each time, a $4$-tuple, i.e., a before-makeup face, a reference face, the after-makeup face by our method and the after-makeup face by one of the baseline, are sent to $20$ participants  to compare.
Note that the two after-makeup faces are  shown in random order.
The participants rate the results into five degrees: ``much better'', ``better'', ``same'', ``worse'', and ``much worse''.
The percentages of each degree are shown in Table \ref{tab:vote_results}.
%
Our method is much better or better than Guo and Sim in  $9.7\%$ and $55.9\%$ cases.
We are much better  than NerualStyle-CC and NerualStyle-CS  in $82.7\%$ and $82.8\%$ cases.


\begin{table}[h]
	\begin{tabular}{|c|c|c|c|c|c|c|}
		\hline
		\hspace{-2.5mm}   \hspace{-2.5mm} &		\hspace{-2.5mm} much better \hspace{-2.5mm} & \hspace{-2.5mm}   better \hspace{-2.5mm} & \hspace{-2.5mm}   same  \hspace{-2.5mm}  &   \hspace{-2.5mm}   worse \hspace{-2.5mm} & \hspace{-2.5mm}   much worse \hspace{-2.5mm} \\
		\hline
		\hspace{-2.5mm} Guo and Sim  \hspace{-2.5mm} & \hspace{-2.5mm} 9.7\% \hspace{-2.5mm} & \hspace{-2.5mm}   55.9\% \hspace{-2.5mm} & \hspace{-2.5mm}   22.4\% \hspace{-2.5mm}
		&  \hspace{-2.5mm}   11.1\% \hspace{-2.5mm} & \hspace{-2.5mm}   1.0\% \hspace{-2.5mm}   \\
		\hline
		\hspace{-2.5mm} NerualStyle-CC  \hspace{-2.5mm} & \hspace{-2.5mm} 82.7\% \hspace{-2.5mm} & \hspace{-2.5mm}   14.0\% \hspace{-2.5mm} & \hspace{-2.5mm}   3.24\% \hspace{-2.5mm}
		&  \hspace{-2.5mm}   0.15\% \hspace{-2.5mm} & \hspace{-2.5mm}   0\% \hspace{-2.5mm}   \\
		\hline
		\hspace{-2.5mm} NerualStyle-CS \hspace{-2.5mm} & \hspace{-2.5mm} 82.8\% \hspace{-2.5mm} & \hspace{-2.5mm}   14.9\% \hspace{-2.5mm} & \hspace{-2.5mm}   2.06\% \hspace{-2.5mm}
		&  \hspace{-2.5mm}   0.29\% \hspace{-2.5mm} & \hspace{-2.5mm}   0\% \hspace{-2.5mm}   \\
		\hline
	\end{tabular}
	\vspace{-4mm}
	\caption{Quantitative comparisons  between  our method and three other makeup transfer methods.   }
	\label{tab:vote_results}
	\vspace{-4mm}
\end{table}

\subsection{More Makeup Transfer Results}
%
In Figure \ref{fig:more_result_same_reference}, for each reference face, we select five most similar looking before-makeup girls.
Then the same makeup is applied on different before-makeup girls.
It shows that the eye shadow, lip gloss and foundation are transferred successfully to the eye lid, lip and face areas.
Note that our method can handle the makeup transfer between different facial expressions.
%
For example, in Figure \ref{fig:more_result_same_reference}, the second girl in the left panel is grinning. However, the reference face is not smiling.
Thanks to the localization property of our method, the lip gloss does not transfer to the teeth in the after-makeup face.

%
In Figure \ref{fig:more_result_same_girl}, for each before-makeup face, we select five most similar looking reference girls.
This function is quite useful in real application, because the users can virtually try different makeup and choose the favorite one.

%
%
%
%
%

%
%

\section{Conclusion}
%
%
In this paper, we propose a novel Deep Localized Makeup Transfer Network   to automatically transfer the makeup from a reference face to a before-makeup face.
The proposed deep transfer network has five nice properties:  with complete function,  cosmetic specific,  localized,  producing naturally looking results and controllable makeup lightness.
Extensive experiments show that our network performs  better than the state-of-the-arts.
In the future, we plan to explore the extensibility of the network.
For example,  one before-makeup face can be combined with two reference faces.
The after-makeup face has the eye shadow of one reference face and lip gloss of another reference face.
%
%

\section*{Acknowledgments}

This work was supported by National Natural Science Foundation of China (No.61572493, Grant U1536203), 100 Talents Programme of The Chinese Academy of Sciences,  and ``Strategic Priority Research Program'' of the Chinese Academy of Sciences (XDA06010701).

%

\small{
\bibliographystyle{named}
\bibliography{ijcai16}
}

\end{document}